\documentclass[10pt,twocolumn,letterpaper]{article}

\usepackage[pagenumbers]{iccv} 

\definecolor{iccvblue}{rgb}{0.21,0.49,0.74}
\usepackage[pagebackref,breaklinks,colorlinks,allcolors=iccvblue]{hyperref}
\usepackage{multirow}
\def\paperID{*****} 
\def\confName{ICCV}
\def\confYear{2025}

\title{REVEAL – Reasoning and Evaluation of Visual Evidence through Aligned Language}

\author{Ipsita Praharaj\( ^1 \)\\
\and
Yukta Butala\( ^1 \)\\
Carnegie Mellon University\( ^1 \)\\
\and
Badrikanath Praharaj\( ^2 \)\\
VIT Bhopal\( ^2 \)\\
\and
Yash Butala\( ^1 \)\\
}

\begin{document}
\maketitle
\begin{abstract}
The rapid advancement of generative models has intensified the challenge of detecting and interpreting visual forgeries, necessitating robust frameworks for image forgery detection while providing reasoning as well as localization. While existing works approach this problem using supervised training for specific manipulation or anomaly detection in the embedding space, generalization across domains remains a challenge. We frame this problem of forgery detection as a prompt-driven visual reasoning task, leveraging the semantic alignment capabilities of large vision-language models. We propose a framework, `REVEAL` (Reasoning and Evaluation of Visual Evidence through Aligned Language), that incorporates generalized guidelines. We propose two tangential approaches - (1) Holistic Scene-level Evaluation that relies on the physics, semantics, perspective, and realism of the image as a whole and (2) Region-wise anomaly detection that splits the image into multiple regions and analyzes each of them. We conduct experiments over datasets from different domains (Photoshop, DeepFake and AIGC editing). We compare the Vision Language Models against competitive baselines and analyze the reasoning provided by them. 
\end{abstract}    
\section{Introduction}
\label{sec:intro}

\begin{figure}
    \centering
    \includegraphics[width=1\linewidth]{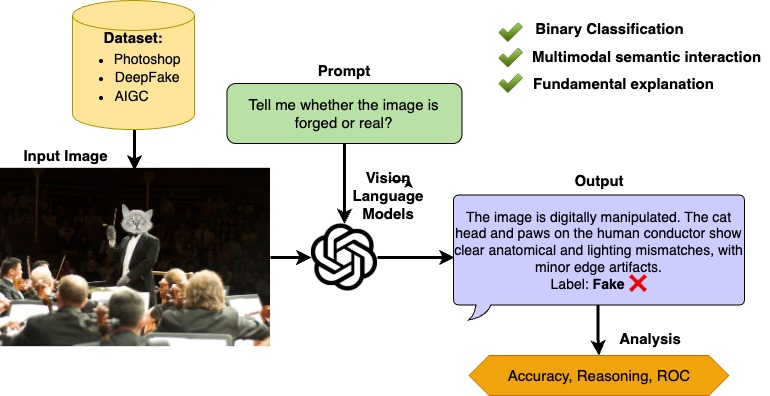}
    \caption{An overview of the image forgery detection task using multimodal large language models. We propose a structured prompt-based approach, REVEAL, to detect forgery and provide nuanced explanations.}
    \label{fig:task-formulation}
\end{figure}

The rapid evolution of deep generative models, such as Stable Diffusion \cite{rombach2021highresolution}, DALLE \cite{marcus2022preliminaryanalysisdalle2, dalle-3} has enabled the creation of high-quality, realistic images. However, these tools pose societal risks—spreading misinformation, enabling cyber-bullying, violating copyright through falsely claimed works \cite{consequences-of-aigc} and also unfair outcomes in creative competitions \cite{art-is-dead}. These concerns underscore the urgent need for reliable image forgery detection to protect authenticity and intellectual property. As shown in \ref{fig:task-formulation}, image forgery detection aim to identify manipulated images and explain the manipulations.

\begin{figure*}[ht]
  \centering
  \begin{subfigure}[b]{0.47\textwidth}
    \includegraphics[width=\linewidth]{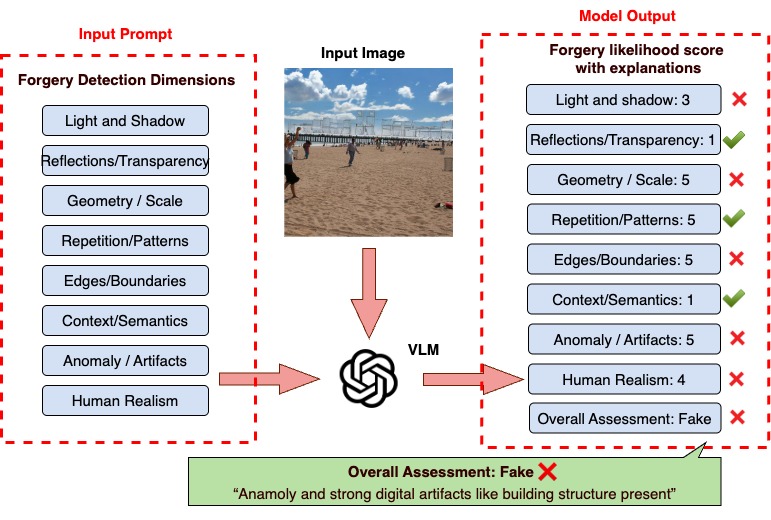}
    \caption{Holistic scene-level evaluation. The model reasons over the entire image for eight different dimensions as shown. It provides the Likert score [1-5] for 'authentic' to 'tampering' for each of the dimensions, followed by an overall assessment.}
    \label{fig:sub1}
  \end{subfigure}
  \hfill
  \begin{subfigure}[b]{0.46\textwidth}
    \includegraphics[width=\linewidth]{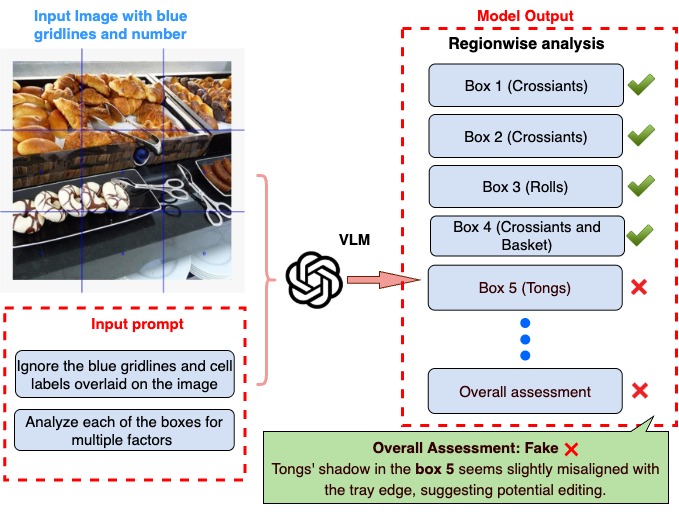}
    \caption{Region-wise anomaly detection: The input image is divided into 9 labeled blocks. The model is expected to reason about forgery detection in each of the boxes before providing the final assessment. }
    \label{fig:sub2}
  \end{subfigure}
  \caption{The images show two kinds of prompt-strategies used in the REVEAL approach}
  \label{fig:reveal-prompts}
\end{figure*}


\noindent\textbf{Increase in Realistic Image Forgeries Using Visual Generative Models:}
Early image generation began with Generative Adversarial Networks (GANs) \cite{goodfellow2014generativeadversarialnetworks}, progressing to more stable variants like SA-GAN \cite{zhang2019selfattentiongenerativeadversarialnetworks} and BigGAN \cite{brock2019biggantraining}. StyleGAN \cite{karras2019stylebasedgeneratorarchitecturegenerative} enabled fine-grained stylistic control, further improving image realism. More recently, diffusion-based models such as Stable Diffusion \cite{rombach2021highresolution}, DALLE-2 \cite{marcus2022preliminaryanalysisdalle2}, and DALLE-3 \cite{dalle-3} have advanced text-guided image synthesis, producing highly realistic forgeries. As a result, each new model increases the challenge of forgery detection.

\noindent\textbf{Multimodal Large Language Models for Forgery Detection:}  
Early models like CLIP \cite{radford2021learningtransferablevisualmodels} and BLIP \cite{li2022blipbootstrappinglanguageimagepretraining} aligned visual and textual features for tasks such as image classification and captioning. Recent MLLMs, including InstructBLIP \cite{dai2023instructblipgeneralpurposevisionlanguagemodels} and LLaVA \cite{liu2023visualinstructiontuningllava}, leverage instruction tuning for complex tasks like visual question answering. Unlike CNN-based approaches \cite{he2015deepresiduallearningimage, delima2020deepfakedetectionusingspatiotemporal, tan2024frequencyawaredeepfakedetectionimproving}, MLLMs provide scene-level understanding and interpretable, text-driven analysis—capabilities critical for robust forgery detection.

\noindent\textbf{Current gaps in Forgery Detection: }
Traditional methods, such as binary classifiers \cite{delima2020deepfakedetectionusingspatiotemporal, guarnera2020deepfakedetectionanalyzingconvolutional} have relied on low-level visual cues, like noise patterns or texture inconsistencies. While effective for specific generative models like GANs \cite{wang2020cnngeneratedimagessurprisinglyeasy}, these approaches struggle with generalizability, particularly against advanced diffusion-based models like Stable Diffusion \cite{rombach2021highresolution} and DALLE-2 \cite{marcus2022preliminaryanalysisdalle2}, due to overfitting on training data \cite{tan2024frequencyawaredeepfakedetectionimproving}.  
To address this, recent MLLM-based approaches like De-Fake\cite{sha2023defakedetectionattributionfake},  AntifakePrompt \cite{chang2024antifakepromptprompttunedvisionlanguagemodels}, ForgeryGPT \cite{liu2025forgerygptmultimodallargelanguage}, and FakeShield \cite{xu2025fakeshieldexplainableimageforgery} integrate visual and linguistic reasoning for interpretable forgery detection and identifying the precise location of the forged detail. 
FakeShield \cite{xu2025fakeshieldexplainableimageforgery} is a fine-tuned MLLM for the forgery classification task trained on real and fake images which also uses visual segmentation mask of the image.
\cite{jia2024chatgptdetectdeepfakesstudy} explored the capability of ChatGPT  in media forensics, highlighting MLLMs’ potential in detecting subtle manipulations. However, their framework is limited to the forensics domain with only GAN generated images.

\noindent\textbf{Key Contributions:}  
We leverage the semantic and spatial reasoning abilities of vision-language models (VLMs) through structured prompting, as illustrated in Figure~\ref{fig:reveal-prompts}. Our work introduces two complementary zero-shot approaches for image forgery detection: (1) \textbf{Holistic Scene Evaluation}, which assesses realism, physics, and semantics at the image level; and (2) \textbf{Region-wise Anomaly Detection}, which analyzes labelled grid segments for localized inconsistencies. While these methods can be further improved with fine-tuning or segmentation pipelines, our focus is on evaluating zero-shot performance with carefully designed instructions. Our main contributions are:

\begin{enumerate}
   \item We propose a dataset-agnostic prompt framework, {REVEAL}, that enables MLLMs to detect forgeries by reasoning from first principles. We evaluate two zero-shot prompting strategies: holistic and region-wise analysis.
    \item We benchmark leading MLLMs against fine-tuned baselines and a simple binary prompt classifier across three image forgery domains from the Fakeshield dataset, and analyze the interpretability and localization of the reasoning produced by {REVEAL}.
\end{enumerate}

\begin{table*}[htbp]
\caption{The table shows the performance of benchmarks and prompt-based baselines over different tasks for image forgery detection. }
\label{tab:detection_performance}
\centering
\resizebox{0.85\textwidth}{!}{%
\begin{tabular}{|l|c|c|c|c|c|c|c|c|c|c|c|c|}
\hline
\multirow{3}{*}{Method} 
& \multicolumn{8}{c|}{\textbf{Photoshop}} 
& \multicolumn{2}{c|}{\textbf{DeepFake}} 
& \multicolumn{2}{c|}{\textbf{AIGC-Editing}} \\
\cline{2-13}
& \multicolumn{2}{c|}{CASIA1+} & \multicolumn{2}{c|}{IMD2020} & \multicolumn{2}{c|}{Columbia} & \multicolumn{2}{c|}{Coverage} & \multicolumn{2}{c|}{} & \multicolumn{2}{c|}{} \\
\cline{2-13}
 & ACC & F1 & ACC & F1 & ACC & F1 & ACC & F1 & ACC & F1 & ACC & F1 \\
\hline
\textbf{Finetuned Baselines} & & & & & & & & & & & &\\
SPAN & 0.60 & 0.44 & 0.70 & 0.81 & 0.87 & 0.93 & 0.24 & 0.39 & 0.78 & 0.78 & 0.47 & 0.05 \\
ManTraNet & 0.52 & 0.68 & 0.75 & 0.85 & 0.95 & 0.97 & 0.95 & 0.97 & 0.50 & 0.67 & 0.50 & 0.67 \\
HiFi-Net & 0.46 & 0.44 & 0.62 & 0.75 & 0.68 & 0.81 & 0.34 & 0.51 & 0.56 & 0.61 & 0.49 & 0.42 \\
PSCC-Net & 0.90 & 0.89 & 0.67 & 0.78 & 0.78 & 0.87 & 0.84 & 0.91 & 0.48 & 0.58 & 0.49 & 0.65 \\
CAT-Net & 0.88 & 0.87 & 0.68 & 0.79 & 0.89 & 0.94 & 0.23 & 0.37 & 0.85 & 0.84 & 0.82 & 0.81 \\
MVSS-Net & 0.62 & 0.76 & 0.75 & 0.85 & 0.94 & 0.97 & 0.65 & 0.79 & 0.84 & 0.91 & 0.44 & 0.24 \\
FakeShield & \textbf{0.95} & \textbf{0.95} & \textbf{0.83} & \textbf{0.90} & \textbf{0.98} & \textbf{0.99} & \textbf{0.97} & \textbf{0.98} & \textbf{0.93} & \textbf{0.93} & \textbf{0.98} & \textbf{0.99} \\
\hline
\textbf{\textcolor{blue}{Baseline Prompt}} & & & & & & & & & & & & \\
LLaVA-1.5-7b-hf & 0.5 & 0 & 0.49 & 0 & 0.5 & 0 & 0.44 & 0 & 0.50 & 0 & 0.50 & 0 \\
Gemini 2.0 Flash & 0.67 & 0.5 & 0.69 & 0.58 & 0.62 & 0.39 & 0.47 & 0.08 & 0.83 & 0.8 & 0.54 & 0.23 \\
GPT-4o & 0.5 & 0.4 & 0.68 & 0.58 & 0.79 & 0.73 & 0.42 & 0 & 0.57 & 0.25 & 0.53 & 0.18\\
GPT-4.1 & 0.83 & 0.8 & \textbf{0.74} & \textbf{0.70} & 0.97 & 0.97 & 0.51 & 0.27 & 0.73 & 0.64 & 0.62 & 0.46 \\
\hline
\textbf{\textcolor{blue}{Holistic Scene level evaluation prompt}} & & & & & & & & & & & & \\
LLaVA-1.5-7b-hf & 0.58 & 0.62 & 0.56 & 0.60 & 0.73 & 0.75 & 0.60 & 0.70 & 0.43 & 0.41 & 0.59 & 0.65 \\
Gemini 2.0 Flash & 0.83 & 0.80 & 0.72 & 0.67 & 0.87 & 0.85 & 0.47 & 0.08 & \textbf{0.93} & \textbf{0.93} & \textbf{0.66} & \textbf{0.53}\\
GPT-4o & 0.75 & 0.67 & 0.69 & 0.62 & 0.94 & 0.94 & 0.49 & 0.26 & 0.62 & 0.39 & 0.55 & 0.26 \\
GPT-4.1 & \textbf{0.92} & \textbf{0.92} & 0.71 & 0.64 & \textbf{0.97} & \textbf{0.97} & 0.47 & 0.08 & 0.66 & 0.47 & 0.62 & 0.42 \\
\hline
\textbf{\textcolor{blue}{Region-wise anomaly detection prompt}} & & & & & & & & & & & & \\
LLaVA-1.5-7b-hf & 0.58 & 0.62 & 0.48 & 0.41 & 0.55 & 0.51 & 0.51 & 0.52 & 0.45 & 0.41 & 0.60 & 0.44 \\
Gemini 2.0 Flash & 0.83 & 0.86 & 0.65 & 0.67 & 0.82 & 0.80 & 0.49 & 0.15 & 0.62 & 0.68 & 0.60 & 0.44 \\
GPT-4o & 0.67 & 0.60 & 0.60 & 0.59 & 0.77 & 0.76 & 0.47 & 0.20 & 0.56 & 0.46 & 0.51 & 0.22\\
GPT-4.1 & \textbf{0.92} & \textbf{0.91} & 0.67 & 0.59 & 0.83 & 0.81 & 0.44 & 0.07 & 0.60 & 0.33 & 0.60 & 0.44 \\
\hline
\end{tabular}
}
\end{table*}

\section{Methodology}

\subsection{Model and Dataset Details}

Similar to the finetuned approach in FakeShield \cite{xu2025fakeshieldexplainableimageforgery}, we select several challenging public benchmark datasets
including PhotoShop tampering (CASIA1+ \cite{CASIA1+}, Columbia \cite{columbia},
IMD2020 \cite{novozamsky2020imd2020}, Coverage \cite{novozamsky2020imd2020}, DeepFake tampering (e.g., FFHQ, FaceApp \cite{dang2020detectiondigitalfacemanipulation-faceapp},
Seq-DeepFake \cite{shao2022detectingrecoveringsequentialdeepfake-seq-deepfake}, and AIGC-Editing data from FakeShield \cite{xu2025fakeshieldexplainableimageforgery}. We select this scope due to the prevalence of these forgery types in real-world scenarios, their semantic complexity, and the availability of established visual cues for benchmarking. We employ open MLLMs, including LLaVA \cite{liu2023visualinstructiontuningllava}, GPT-4o \cite{openai2024gpt4ocard}, and GPT-4.1 \cite{gpt-4}, Gemini \cite{gemini-2}.

\subsection{Prompting Strategies}

We systematically evaluate on the baseline prompts and two prompts specialized prompts (provided in Appendix~\footnote{We upload the appendix to Google Drive to adhere to the page limit. \href{https://drive.google.com/file/d/11v5vKDiXWYZ1Akx8QabzYIrONb6xYhQm/view?usp=sharing}{Link}}, each targeting a different aspect of VLM reasoning:

\begin{itemize}
    \item \textbf{Baseline Prompt:} A simple binary classification prompt (“Is this image real or fake?”) to assess the VLM’s out-of-the-box ability without explicit reasoning or spatial cues.
    \item \textbf{Holistic Scene-Level Prompt (Image Semantics):} A detailed checklist-based prompt requiring the model to assess global scene plausibility across eight forensic dimensions: \emph{lighting/shadow, reflections/transparency, perspective/geometry, repetition/patterns, edge/boundary, contextual/semantic consistency, anomaly/artifact detection, and human/object realism} based on~\cite{hohloch2024homoclinicfloerhomologydirect}. 
    For each factor, the model provides a Likert score from 1 to 5 (authentic to forged) and a concise factor-wise justification, followed by a global label and a brief major reasoning of the most decisive evidence from the 8 factors. 
    \item \textbf{Region-wise Anomaly Detection Prompt (Image Detailing):} Inspired by recent advances in visual grounding - Set of Mark~\cite{yang2023setofmarkpromptingunleashesextraordinary}, this prompt overlays a 3x3 labelled grid over the image and instructs the model to first perform a holistic assessment, then analyze each grid cell for local anomalies only if no holistic cues are found. The model integrates both overall and localized (cell-specific) findings into a single, concise explanation, referencing cell numbers for local issues.
\end{itemize}


 
\begin{figure}
    \centering
    \includegraphics[width=1\linewidth]{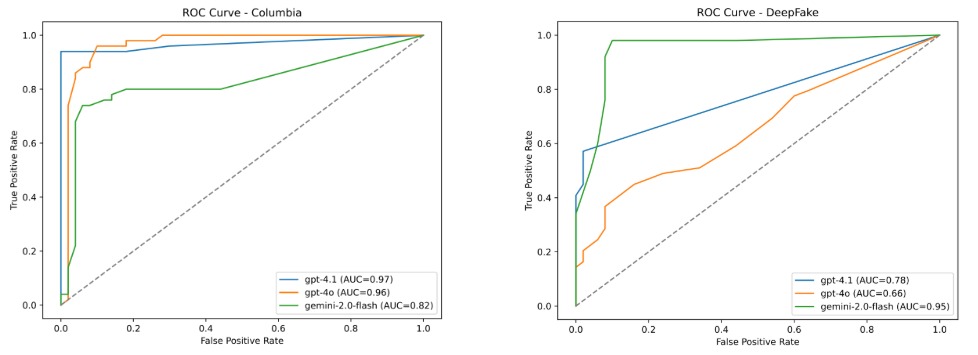}
    \caption{We plot the ROC curve using average of factor-scores for the Holistic-eval approach as a proxy to the probability. We find that GPT-4.1 performs well over the Columbia dataset while Gemini is the best amongst the baselines over DeepFake dataset.  }
    \label{fig:roc}
\end{figure}

\begin{figure*}[htbp]
    \centering
    \includegraphics[width=0.95\textwidth]{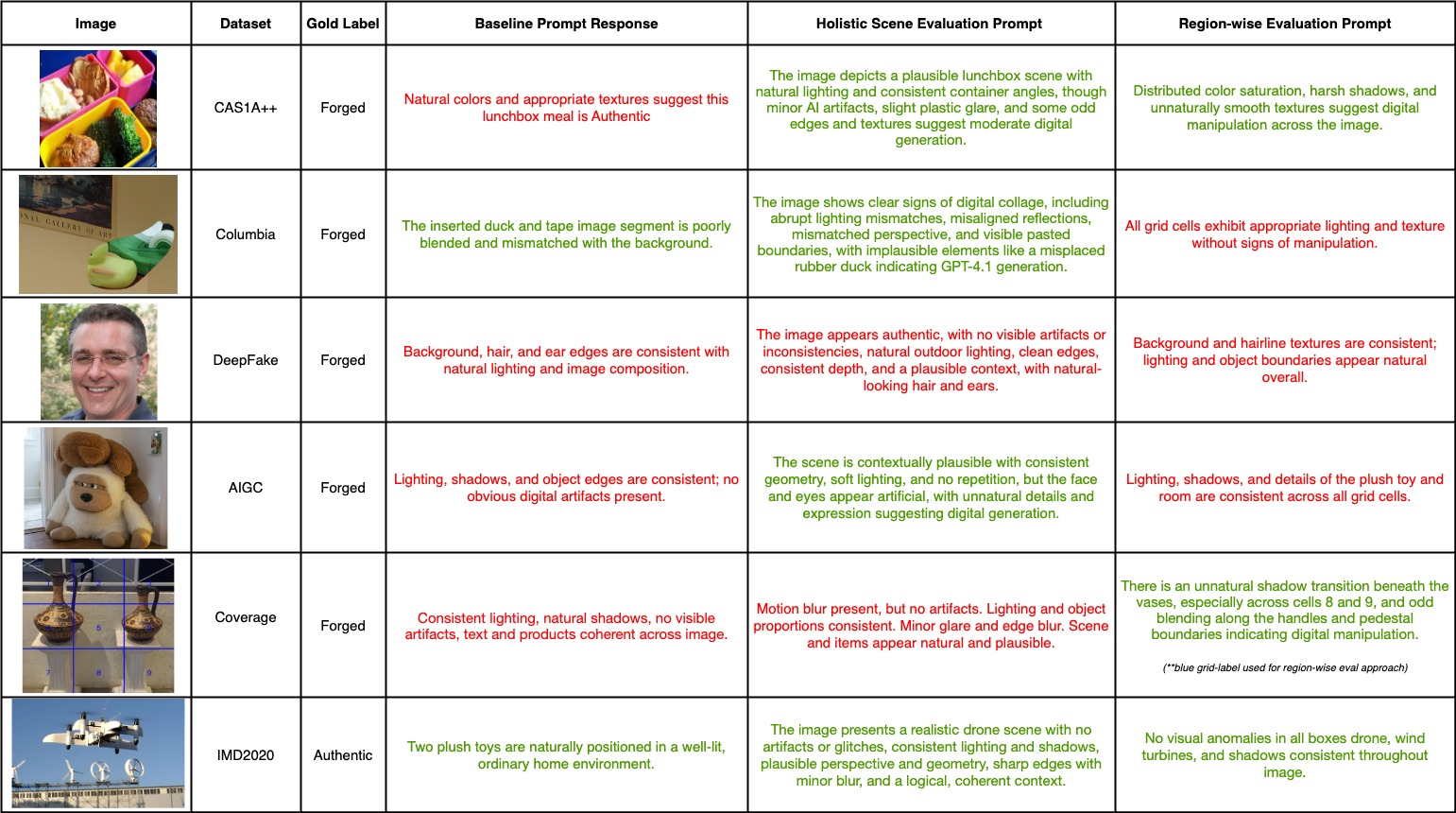}
    \caption{We pick examples for qualitative analysis. The outputs for "Holistic Scene-Evaluation" and "Region-wise Evaluation" prompt are made concise using GPT.}
    \label{fig:qual_anal}
\end{figure*}

\subsection{Metrics}
\label{metrics}
Using the predicted labels from each of the approaches (baseline, holistic, region-wise), we report accuracy and F1 score over the classification task of image forgery detection. 
Additionally, for the Holistic Scene-level evaluation, we average the Likert scores for each of the factors to get a tampering score between 0 to 1. It is used to obtain the ROC curve for threshold-independent comparison between different MLLMs, as shown in figure \ref{fig:roc}.

\section{Results}

\subsection{Quantitative Results}

The main findings as shown in the table \ref{tab:detection_performance} are:

\begin{itemize}
    \item Structured prompts (holistic and region-wise) in general significantly improve the performance over baseline prompts.   For example, GPT-4.1's accuracy on CASIA1+ increases from 0.83 (baseline) to 0.92 (holistic/region-wise), and Gemini's F1 on Columbia \cite{columbia} rises from 0.39 (baseline) to 0.85 (holistic) and 0.80 (region-wise).
    \item Structured prompts have the greatest impact on datasets where the baseline model struggles—typically those with subtle, complex, or localized manipulations (e.g., DeepFakes, AIGC-Editing). For example, Gemini and GPT-4o show significant improvement on DeepFake and AIGC datasets when using structured prompts, as these prompts help the models focus on specific forensic cues. 
    Similarly, for the smaller model LLaVA, structured prompts improve performance consistently and significantly, where accuracy and F1 scores rise from near-zero to moderate, showing LLaVA can use scene-level cues when prompted with explicit guidance. Thus, structured prompting is most beneficial for challenging datasets and weaker models that can follow the prompt. However, in contrast, on datasets like Photoshop forgeries (CASIA1+, Columbia), where GPT-4.1 already performs well, the benefit of structured prompts is less pronounced.  
    \item Prompt-based methods achieve low F1-score over the Coverage \cite{wen2016coverage} data due to subtle and distributed edits.

\end{itemize}

\subsection{Qualitative Analysis}

Figure~\ref{fig:qual_anal} shows sample GPT-4.1 results. Key findings:

\begin{itemize}
    \item \textbf{Region-wise prompts excel at local manipulations:} Grid-based prompts effectively detect localized edits (e.g., AIGC-Editing, Photoshop splicing), but may miss manipulations spanning multiple cells if inter-cell consistency is not considered. For example, in Figure~\ref{fig:qual_anal} (row 2), the model finds each cell authentic but misses the global inconsistency.
    \item \textbf{Labelled grids outperform imagined grids:} Aligning with the observation in \cite{yang2023setofmarkpromptingunleashesextraordinary}, overlaying grids and numbers on images significantly improved detection compared to asking the model to imagine the regions.
    \item \textbf{Holistic prompts capture global cues:} Holistic prompts are best for scene-wide manipulations (e.g., lighting, context), but struggle with localized edits. Example: in Row 3 (DeepFake) and Row 5 (Coverage) the model fails to detect forgery.
    \item \textbf{Factor-label correlation aids interpretability:} Likert scores for each of the factors help deduce the cues that matter for each dataset. We find that Human/Object Realism and Anomaly/ Artifacts are most important for AIGC Editing and DeepFakes. Edges/Boundaries and Reflections are key to detecting Photoshop Forgeries. 
    \item     \textbf{ROC curve using Likert scores:} The AUC score is less affected by imbalanced data and is helpful for threshold agnostic comparison between MLLMs.
    Figure ~\ref{fig:roc} plots the ROC curves using the Likert scores as explained in section \ref{metrics}. They show quantified differences in various MLLM's classification. performance 
\end{itemize}


\section{Conclusion}

We introduced REVEAL, a prompt-driven framework for image forgery detection with vision-language models. By leveraging holistic and region-wise evaluation, REVEAL enables interpretable, zero-shot reasoning across diverse forgery types. Our experiments show that structured prompts substantially improve both detection accuracy and explanation quality over baseline prompts. As generative models advance, prompt-based methods offer a scalable alternative to fine-tuning, adapting quickly to new manipulation techniques. Future work will integrate REVEAL with segmentation models to further enhance localization and region-specific analysis.
{
    \small
    \bibliographystyle{ieeenat_fullname}
    \bibliography{main}
}

\end{document}